# Equilibrio de Carga para Transformadores de Distribución Eléctrica Mejorando la Calidad de Servicio en Fin de Línea

*Juan M. Bordón*
*Departamento de Matemática, Universidad Nacional de Tucumán, Facultad de Ciencias Exactas y Tecnología*
*jmbordon@herrera.unt.edu.ar*

*Victor A. Jimenez*
*Grupo de Investigación en Tecnologías Informáticas Avanzadas, Universidad Tecnológica Nacional - Facultad Regional Tucumán*
*victoradrian.jimenez@frt.utn.edu.ar*

*Adrian Will*
*Grupo de Investigación en Tecnologías Informáticas Avanzadas, Universidad Tecnológica Nacional - Facultad Regional Tucumán*
*adrian.will@gitia.org*

## Resumen

*La distribución de energía eléctrica enfrenta desafíos globales, como la creciente demanda, la integración de generación distribuida, las pérdidas elevadas y la necesidad de mejorar la calidad del servicio. En particular, el desbalance de cargas, donde las cargas no están distribuidas uniformemente entre las fases de los circuitos, puede reducir la eficiencia, acortar la vida útil de los equipos y aumentar la susceptibilidad a interrupciones del servicio. Los métodos que implican mover cargas de una fase a otra pueden ser costosos, pero son efectivos cuando se dispone de medidores inteligentes y se llevan a cabo de manera eficiente. Este trabajo propone el uso de algoritmos genéticos para identificar de manera óptima las cargas a redistribuir, con el fin de mejorar tanto el balance de cargas como la calidad de la tensión en los nodos finales de la red, minimizando la cantidad de cambios necesarios. El algoritmo fue evaluado mediante simulaciones utilizando PandaPower, una herramienta de análisis de flujo de carga, modelando redes simples basadas en características reales del sistema eléctrico en Tucumán.*

## Introducción

La distribución de energía eléctrica es uno de los mayores desafíos que enfrentan las empresas de servicios públicos a nivel mundial. La creciente demanda, los altos niveles de pérdidas, las fallas y cortes del servicio, así como la necesidad de mejorar la calidad del servicio para los consumidores, son solo algunas de las problemáticas relacionadas [14]. Entre ellas, el desbalance de cargas genera un flujo de corriente desigual en las líneas eléctricas, lo que provoca una caída de voltaje excesiva, reduce la eficiencia en la entrega de energía eléctrica, acorta la vida útil de los equipos eléctricos, aumenta la susceptibilidad a interrupciones del suministro eléctrico y disminuye la calidad del servicio para los usuarios [15]. Este problema se presenta generalmente cuando las cargas eléctricas no están distribuidas de manera equilibrada entre las fases del transformador [6].

Existen varios métodos para balancear las cargas en las redes de distribución eléctrica. Uno de ellos es realizar una buena planificación y diseño de la red previo a su implementación, optimizando la ubicación de transformadores y la distribución de cargas para lograr un balance uniforme. En redes ya operativas, se emplean dispositivos como bancos de capacitores [13] o filtros armónicos para reducir caídas de tensión y mejorar ligeramente el desequilibrio, aunque estos métodos son costosos y no siempre ofrecen soluciones duraderas. Los métodos de redistribución de cargas, por otro lado, implican reubicar las cargas eléctricas entre fases para balancearlas. La efectividad de cada método varía según las características de la red y las cargas conectadas. Sin embargo, este último enfoque es especialmente eficiente cuando se dispone de infraestructura de medición avanzada que permita monitorear el consumo de los clientes, ya que optimiza los costos de implementación, mejora la eficiencia y ofrece soluciones sostenibles en el tiempo. Para su aplicación, es necesario utilizar un algoritmo de búsqueda que identifique los cambios requeridos y ejecutar dichas modificaciones en campo.

Se han propuesto numerosos algoritmos para el balance de cargas, incluyendo soluciones analíticas [17], optimización combinatoria [8] y lógica difusa [2], entre otros [15, 3]. Los algoritmos más comunes se basan en optimización heurística, adaptando las funciones objetivo según las necesidades específicas. Se han utilizado algoritmos inspirados en el Sistema Inmune para minimizar la corriente de neutro y los costos de reconexión de fases [10], y la Optimización por Enjambre de Partículas para reducir el desequilibrio de corrientes [20]. Los algoritmos genéticos también son ampliamente aplicados, aunque suelen usar arquitecturas tradicionales [9], con funciones objetivo que penalizan los cambios de fase de diversas maneras. La elección del algoritmo depende de los datos disponibles y las condiciones operativas específicas de cada escenario. En [11], se propone un método basado en *Deterministic Crowding*, una variante de algoritmos genéticos tipo *Niching*, que se destaca por el uso de datos de fácil acceso sin necesidad de medir todos los clientes, y su capacidad para evitar cambios de fase innecesarios para reducir costos.

Desde el punto de vista operativo, reconectar clientes para balancear las cargas presenta un desafío significativo. Varias combinaciones de reconexiones pueden ofrecer mejoras similares en el balance de cargas en el transformador, pero estos cambios afectan considerablemente la tensión aguas abajo. Por lo tanto, al seleccionar los cambios a realizar, es crucial considerar los valores de voltaje resultantes, especialmente en los nodos finales de la red. Este aspecto es fundamental para la calidad del servicio, un parámetro clave para las empresas distribuidoras.

En este trabajo proponemos el uso de algoritmos genéticos, utilizando datos de medidores inteligentes, para identificar qué clientes deben ser reconectados y a qué fase, con el fin de balancear las cargas en subestaciones transformadoras. Se emplea la variante *Deterministic Crowding*, pero, a diferencia de estudios previos, se añadirá como criterio adicional de búsqueda mejorar la calidad de la tensión en los nodos finales de la red. De este modo, las soluciones que se encuentren permitirán balancear las cargas sin descuidar los niveles de tensión en cada fase dentro de los márgenes recomendados especialmente en los nodos finales de la red.

Este trabajo se organiza de la siguiente manera: La Sección 2 describe en detalle el algoritmo y la metodología utilizada para encontrar los cambios necesarios para balancear las cargas y mantener el nivel de tensión dentro de los límites deseados; La Sección 3 describe los escenarios de prueba y los resultados obtenidos para validar la propuesta; Finalmente, la Sección 4 presenta las conclusiones y se describen los trabajos futuros.

## Algoritmo para equilibrar las cargas y reducir caída de tensión

Antes de implementar cualquier algoritmo y hacer efectivos los cambios en la red eléctrica, es esencial realizar simulaciones exhaustivas para evaluar el comportamiento de la red bajo diversas condiciones y garantizar que las modificaciones no comprometan la estabilidad ni la eficiencia del sistema. En el caso del desequilibrio de cargas, es necesario simular los cambios de fase de los clientes seleccionados por el algoritmo de búsqueda. En trabajos previos [11], este proceso se abordó simplemente sumando y restando el consumo de los clientes movidos entre fases, lo cual es adecuado cuando sólo se requiere balancear la corriente en el transformador. Sin embargo, en este trabajo requerimos usar valores de caída de tensión que dependen de la topología de la red. Por este motivo, se requiere una simulación más compleja mediante un análisis de flujo de potencia (power flow), también conocido como análisis de flujo de carga (load flow analysis). Esta herramienta es fundamental en el estudio de sistemas eléctricos de potencia, permitiendo determinar tensiones en cada nodo, corrientes, pérdidas en las líneas y la energía total suministrada por la subestación transformadora. Este análisis es estático y se centra en el estado estacionario del sistema, es decir, en condiciones de equilibrio de todas las variables eléctricas. Para este trabajo, empleamos *PandaPower* [19], una herramienta de código abierto en Python diseñada para el análisis de sistemas de potencia, que ofrece flexibilidad, facilidad de uso y una detallada biblioteca de modelos de componentes de redes eléctricas.

Para implementar la simulación, es necesario contar con un modelo detallado de la red eléctrica que respete su topología, incluyendo circuitos, tipo y longitud del cableado, así como la distribución geográfica de los clientes (ver Figura 1). También se deben disponer de mediciones de consumo eléctrico proporcionadas por los medidores inteligentes, tanto de los clientes como de la subestación transformadora. Estas mediciones, generalmente tomadas cada 15 o 60 minutos, permiten obtener un conjunto de datos lo suficientemente extenso como para evaluar si los cambios sugeridos por el algoritmo mejoran el desbalance de cargas y la caída de tensión a lo largo del tiempo. El algoritmo para determinar los cambios de fase necesarios debe realizar múltiples simulaciones para probar distintas combinaciones. Los métodos heurísticos de búsqueda aleatoria, como los Algoritmos Genéticos, son ideales para esta tarea debido a la complejidad del problema y que, a diferencia de otros métodos, pueden obtener buenos resultados en tiempos razonables. Además, una solución que respete las restricciones operativas suele ser más efectiva y fácil de implementar que una solución determinística que pueda resultar costosa computacionalmente o impráctica.

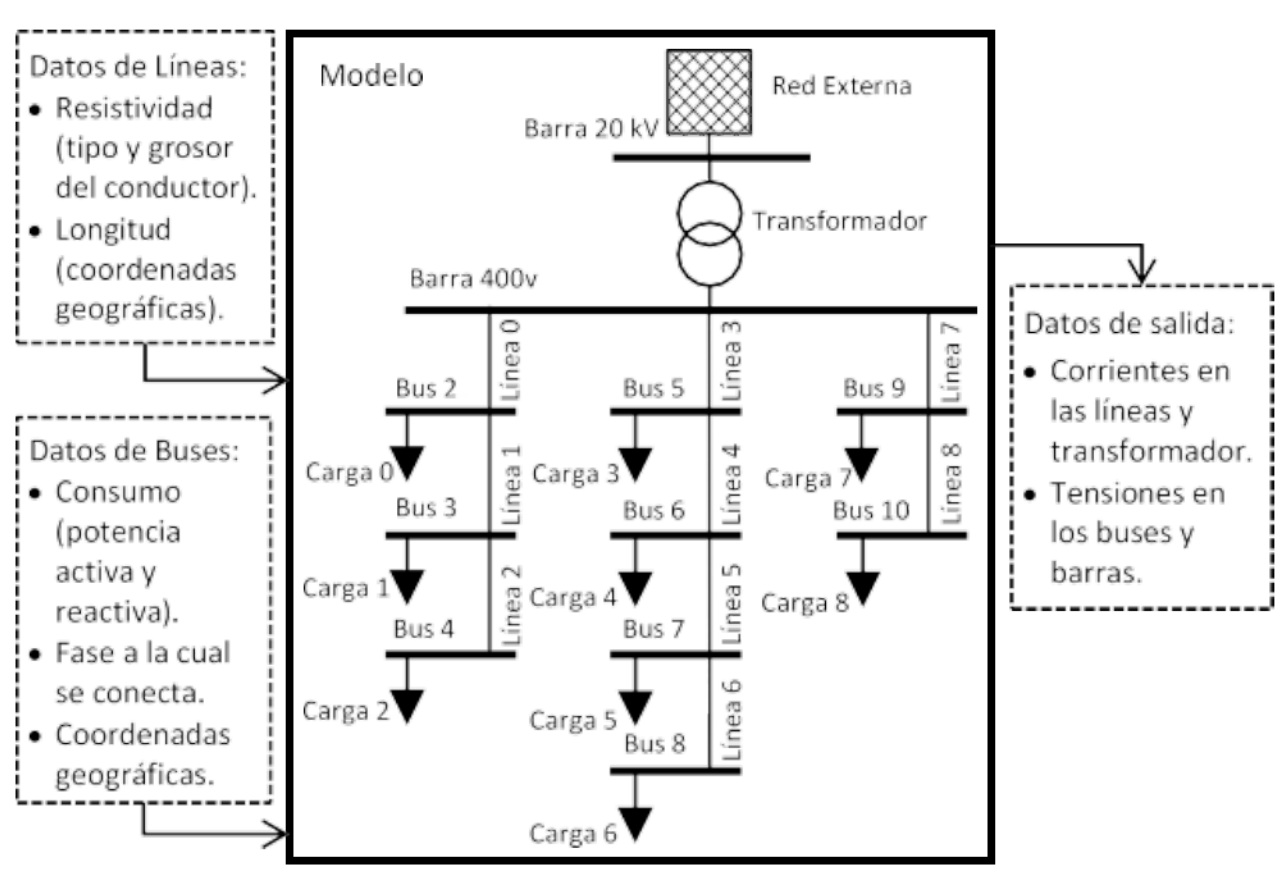

**Figura 1: Modelado de una red eléctrica utilizado para realizar un análisis de flujo de carga.**

## Algoritmos Genéticos

Entre las heurísticas bio-inspiradas, destacan los Algoritmos Genéticos (AG), basados en la teoría de la evolución de Darwin, que buscan emular el proceso evolutivo a través de la selección natural [5]. En la naturaleza, los individuos mejor adaptados tienen mayores posibilidades de supervivencia y reproducción, transmitiendo sus características a las generaciones futuras. Este proceso de selección natural impulsa la evolución y mejora continua de las especies. En los AG, los individuos representan posibles soluciones al problema en cuestión, y su calidad se evalúa mediante una función de aptitud o *fitness*. Esta función actúa como el entorno selectivo en la naturaleza, determinando qué tan bien se adapta cada solución a la tarea específica.

Existen diversas variantes de AG, que utilizan distintos operadores genéticos de formas variadas. En este trabajo, emplearemos una variante de tipo *Niching* [16], diseñada para situaciones que requieren múltiples soluciones al problema. Estos algoritmos permiten encontrar y preservar todos los óptimos locales en el espacio de búsqueda. En particular, utilizaremos el algoritmo Deterministic Crowding (DC) [7], cuyo pseudocódigo se muestra en la Figura 2. El algoritmo parte con una población de individuos generados al azar. No tiene operador de selección, por lo que el primer paso en cada iteración es obtener una descendencia a partir de la población principal mediante el operador de *cruzamiento*, participando toda la población en este paso. Para garantizar la aleatoriedad, la población de padres se mezcla y se agrupa en pares al azar. Luego, los hijos se evalúan con la función *fitness*. Cada hijo compite con el padre más cercano según una métrica de distancia *D* (descrita en la Sección 2.1.2), ganando el de mejor fitness, lo que asegura que solo compitan individuos dentro del mismo nicho. La condición de terminación del algoritmo, común en AG y otras heurísticas, es que el algoritmo debe esperar una cantidad preestablecida de generaciones sin encontrar una mejor solución que la obtenida hasta el momento. Esto permite al algoritmo ajustarse al problema sin añadir procesamiento excesivo ni parámetros adicionales que el usuario final deba ajustar.

El algoritmo DC ofrece varias ventajas. En primer lugar, el número reducido de parámetros simplifica su ajuste. Además, una vez establecidos, los parámetros han demostrado ser estable a repeticiones. Esto significa que los resultados obtenidos presentan poca variación entre diferentes ejecuciones, tanto en la calidad de las soluciones (valor de fitness constante) como en las soluciones mismas (muy similares entre distintas pruebas). Aunque el objetivo principal de este trabajo no es encontrar múltiples soluciones, se seleccionó el algoritmo DC por dichas ventajas y porque ha sido efectivo en problemas de complejidad similar, produciendo consistentemente buenos resultados con una cantidad razonable de procesamiento [11].

```
Entrada:
- tam_p: cantidad de individuos en la población
- tam_m: cantidad de individuos mutados
- prob_mut: probabilidad de mutar gen
/* población inicial con individuos aleatorios */
poblacion = generar_poblacion_aleatoria(tam_p)
mientras comprobar_condicion_terminacion() = falso hacer
    /* fase de cruzamiento */
    mezclar(poblacion)
    descendencia = {}
    para k = 1 ... tam_p con paso 2 hacer
        hijos = cruzar(poblacion[k], poblacion[k+1])
        concatenar(descendencia,hijos)
    /* calcular fitness de individuos */
    para k = 1 ... tam_p hacer
        evaluar(descendencia[k])
    /* comparación y sustitución */
    para k = 1 ... tam_p con paso 2 hacer
        p1 = poblacion[k]
        p2 = poblacion[k+1]
        c1 = descendencia[k]
        c2 = descendencia[k+1]
        si D(p1,c1) + D(p2,c2) ≤ D(p1,c2) + D(p2,c1) entonces
            poblacion[k] = competencia(p1, c1)
            poblacion[k+1] = competencia(p2, c2)
        en otro caso
            poblacion[k] = competencia(p1, c2)
            poblacion[k+1] = competencia(p2, c1)
solucion = extraer_mejor(poblacion)
```

**Figura 2: Pseudocódigo de Deterministic Crowding.**

## Codificación de las soluciones

Al aplicar el algoritmo DC para resolver el problema de balance de carga, es fundamental definir la codificación de las soluciones, es decir, cómo se representarán computacionalmente para su procesamiento. En este caso, se elige la codificación entera, la cual es eficiente para valores discretos. Cada cliente se representa con un valor de 1, 2, 3, que corresponde a las fases A, B o C de la subestación transformadora, respectivamente. De este modo, cuando el valor asignado a un cliente difiere de la fase a la que está realmente conectado, se deduce que éste debe cambiar de fase. En cambio, si el valor asignado coincide con la fase real del cliente, no se requiere ningún cambio. La Figura 3 muestra a modo de ejemplo una solución para una red eléctrica con 8 clientes en total. Aquellos que tengan conexión trifásica o no tengan un medidor de consumo eléctrico disponible (normalmente por problemas de comunicación del medidor) no se tendrán en cuenta en las soluciones.

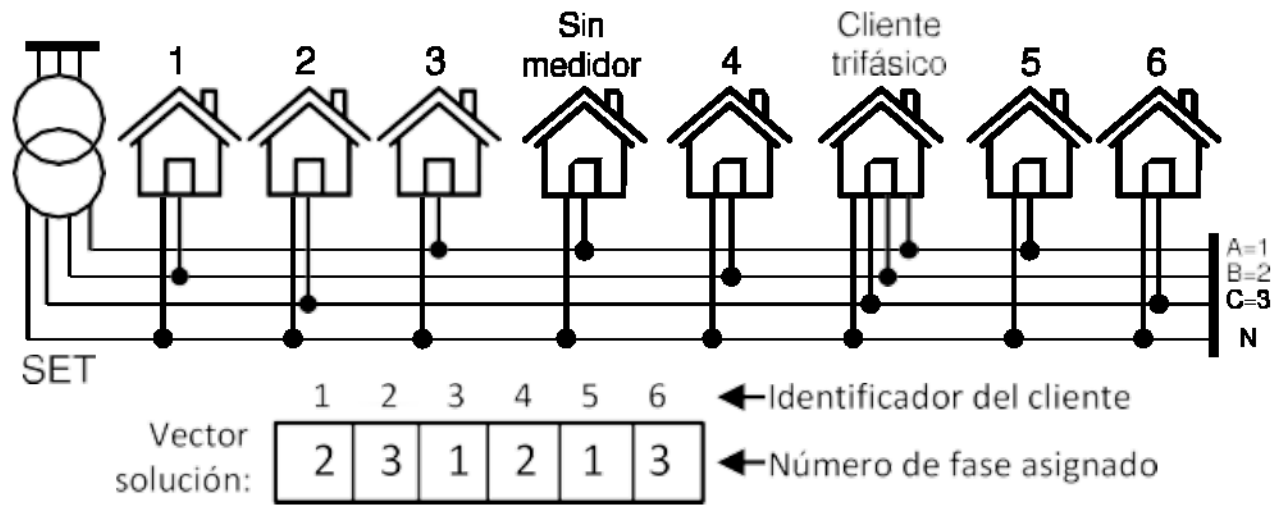

**Figura 3: Codificación de los vectores soluciones para el problema de balance de cargas.**

Se asume que se conoce con precisión la conexión de cada cliente al transformador y su fase correspondiente. Aunque, en la práctica, esta información puede ser imprecisa o incompleta, se puede obtener a partir de mediciones de consumo eléctrico, empleando métodos preexistentes con un alto nivel de precisión [12].

### Operadores Genéticos

Para utilizar el algoritmo DC, es necesario definir el operador de cruzamiento y la métrica de distancia entre soluciones (ya que esta variante de AG no utiliza operadores de selección ni de mutación). En cuanto al operador de cruzamiento, se emplea el *Uniform Crossover*. Este método intercambia los valores de cada gen entre dos padres con una probabilidad del 0,5 , aplicándolo a toda la población en cada generación. Este operador es eficaz para explorar el espacio de búsqueda, ya que permite que cada gen de los padres pueda ser intercambiado. Finalmente, la métrica de distancia en el espacio de soluciones utilizada (denominada función $D$ en el pseudocódigo de la Figura 2) es la distancia *Hamming*, que se calcula como la cantidad de elementos diferentes entre dos vectores. Esta métrica es adecuada para la codificación entera utilizada en el problema de balance de cargas.

### Función objetivo

La función objetivo o función *fitness* requiere mediciones del consumo de los clientes tomadas durante un período suficientemente largo para garantizar que la nueva configuración de fases mantenga el balance a lo largo del tiempo. Estos datos se utilizan para simular los cambios de fase mediante un análisis de flujo de carga con el software PandaPower, como se describió anteriormente. A partir de los resultados de la simulación, el fitness se calculará utilizando el índice de desbalance y la caída de tensión, que se describen a continuación.

### Índice de desbalance de cargas

Existen diversas normas y estándares que definen diferentes métodos para calcular e interpretar el desbalance de cargas en una red eléctrica trifásica [9]. En este trabajo, utilizamos la norma NEMA MG-1 [18], que propone un cálculo sencillo del índice de desbalance $b_t$ mediante la Ecuación 1, donde $I_t^{\max}$ es la magnitud de la corriente de fase más grande, y $I_t^{\text{prom}}$ es el promedio de las magnitudes de las tres corrientes de fases.

$$b_t = 100 \cdot \frac{I_t^{\max} - I_t^{\text{prom}}}{I_t^{\text{prom}}} \tag{1}$$

El índice $b_t$ es una métrica eficiente para medir el desbalance, permitiendo su aplicación con solo la magnitud de las corrientes de fase. Sin embargo, sólo cuantifica el desbalance en un instante específico. Para evaluar el desbalance a lo largo de un período extendido, utilizamos el índice $B$ [11] definido por la Ecuación 2, donde $n$ es la cantidad de muestras de datos consideradas. El índice $B$ se expresa en las mismas unidades que el índice original, por lo que su interpretación es similar.

$$B = \sqrt{\sum_{t=0}^{n} \frac{b_t^2}{n}} \tag{2}$$

Según la norma NEMA MG-1, el límite máximo aceptado para el desbalance de tensión es del 1 %, lo que corresponde a un desbalance máximo de 20 % para la corriente [1]. Sin embargo, en este trabajo adoptamos un valor deseado del 15 %, en concordancia con las recomendaciones operativas de la empresa distribuidora eléctrica en la provincia de Tucumán.

### Caída de tensión en fin de línea

La caída de tensión se refiere a la pérdida de potencial eléctrico cuando la corriente fluye a través de un conductor. Esta pérdida es una manifestación directa de la resistividad eléctrica del material y está relacionada con el efecto Joule, ya que la energía perdida se convierte en calor. La caída de tensión en un cable es directamente proporcional a su resistividad y longitud, e inversamente proporcional al área de su sección transversal. La Figura 4 muestra una línea de transmisión con varios clientes conectados y cómo la tensión disminuye debido al cable y a la carga de los clientes. En una red trifásica, la distribución desigual de las cargas entre las fases también afecta la tensión a lo largo de la línea. Estos y otros factores combinados complican la estimación precisa de la caída de tensión y el impacto de los cambios de fase en los clientes, lo que hace necesaria una simulación para obtener resultados fiables.

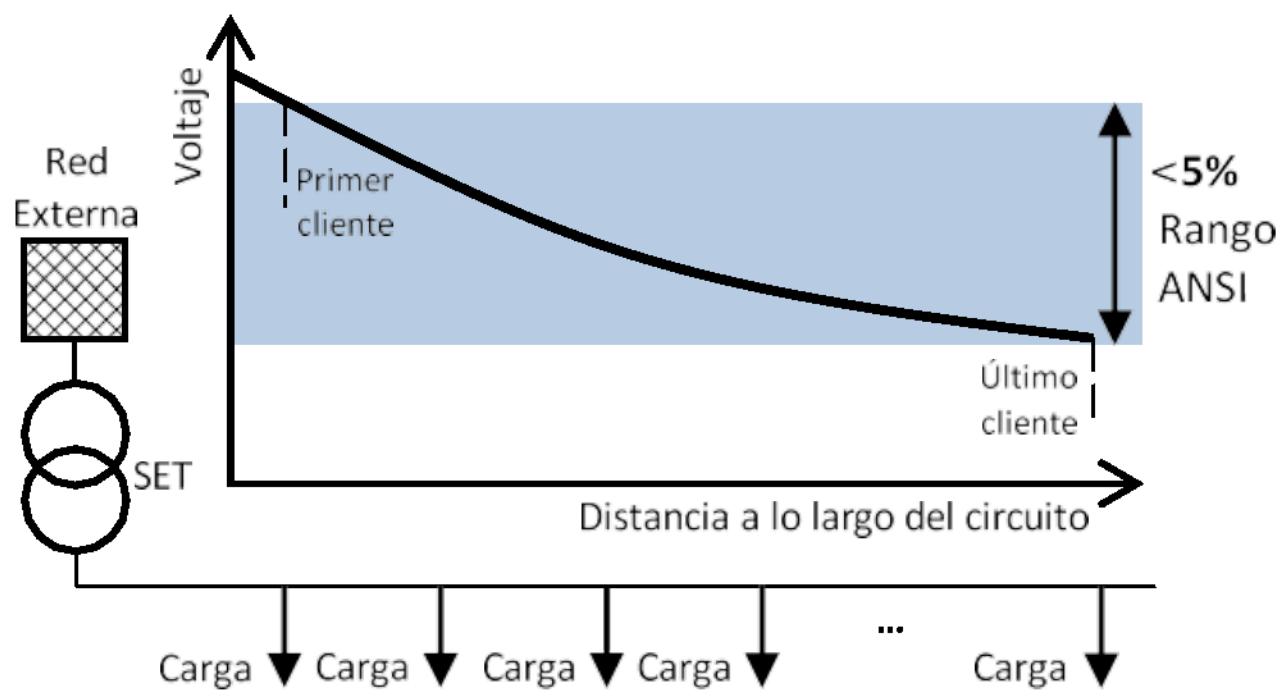

**Figura 4: Esquema de la caída de tensión debido a diferentes servicios conectados a lo largo de una línea de distribución de baja tensión.**

Las empresas de distribución suelen establecer un umbral de caída de tensión, que define el nivel máximo permitido antes de que se considere inaceptable o que pueda afectar negativamente el funcionamiento de los equipos conectados a la red. La caída de tensión en el final de línea, $\Delta V_t$ en un tiempo $t$, se calcula mediante la Ecuación 3. Este porcentaje se obtiene comparando la tensión del último nodo del circuito, $V_t^f$, con el valor nominal del sistema $V_t^0$ registrado en la subestación transformadora (230.9v, correspondiente a una tensión fase-fase del secundario de 400v). El umbral que adoptamos para la caída de tensión es 5 %.

$$\Delta V_t = 100 \cdot \frac{V_t^0 - V_t^f}{V_t^0} \tag{3}$$

En este trabajo, al analizar la red eléctrica durante un periodo determinado por $n$ muestras tomadas a lo largo del tiempo, el valor de la caída de tensión para dicho periodo se calcula utilizando la Ecuación 4. Este valor representa el máximo registrado de caída de tensión durante el periodo de análisis, y se considerará sólo la fase que tenga mayor caída de tensión.

$$\Delta V = \text{máx}\,\{\Delta V_t\}_{t=1}^{n} \tag{4}$$

## Cantidad de cambios

Cuando la reasignación de fase de los clientes se realiza manualmente, es crucial considerar soluciones que propongan la mínima cantidad posible de cambios debido a la complejidad, el tiempo, los riesgos, la incomodidad de los usuarios y los costos monetarios asociados al proceso. Dada una solución $S$ propuesta por el algoritmo de optimización, la cantidad de cambios necesarios $N$ se calcula mediante la Ecuación 5, donde $m$ es la cantidad de clientes considerados en el espacio de búsqueda, $p_i \in \{1, 2, 3\}$ es la fase inicial del cliente $i$, $s_i \in \{1, 2, 3\}$ es la fase propuesta para el cliente $i$, y $c_i$ toma el valor 1 si la solución $S$ cambia la fase del cliente $i$ o 0 en caso contrario.

$$N = \sum_{i}^{m} c_i, \text{ donde } c_i = \begin{cases} 0 & \text{si } s_i = p_i \\ 1 & \text{si } s_i \neq p_i \end{cases} \tag{5}$$

## Combinación de objetivos

El valor de fitness de cada individuo de la población se obtiene utilizando la función definida por la Ecuación 6.

$$f(S) = 1 - \frac{\alpha B/B^{\max} + \beta \Delta V/\Delta V^{\max} + \gamma N/N^{\max}}{\alpha + \beta + \gamma} \tag{6}$$

Donde:

- $B$: índice de desbalance calculado con la Ecuación 2 para el periodo analizado.
- $\Delta V$: caída de tensión producida en el nodo final del circuito obtenida con la Ecuación 4 para el periodo analizado.
- $N$: cantidad de cambios necesarios dada por la solución obtenida mediante la Ecuación 5.
- Constantes $B^{\max}$, $\Delta V^{\max}$ y $N^{\max}$: son los valores máximos que se espera para cada índice utilizado, los cuales determinan el rango donde se mueven cada uno de los valores objetivos a optimizar. En nuestro caso adoptamos respectivamente los valores 100, 10 y 50.
- Parámetros $\alpha$, $\beta$ y $\gamma$: permiten controlar a cual objetivo se le da más importancia. Al dividir el total en la suma de estos parámetros, la función de fitness así definida se convierte en una combinación convexa de los objetivos, permitiendo ajustar fácilmente sus valores.

El AG deberá entonces buscar el máximo de esta función. Alternativamente, se podría plantear una optimización utilizando un Algoritmo Evolutivo Multi-Objetivo de Frente de Pareto [5], aunque en ese caso también sería necesario tomar una decisión sobre la importancia relativa de cada objetivo al seleccionar una solución entre las posibles. Por esta razón, decidimos asignar a priori los valores de los pesos de cada objetivo en función de las condiciones operativas y aplicar una optimización mono-objetivo.

## Validación del algoritmo

Para evaluar el comportamiento del algoritmo se utiliza un escenario de prueba simple con un modelo de red eléctrica conformada por un transformador que alimenta a un grupo de clientes conectados secuencialmente en un único circuito, como se muestra en la Figura 5. Para modelar esta red utilizando Panda-Power, se consideraron las características de redes eléctricas típicas en barrios residenciales de Tucumán [4]:

- Subestación Transformadora de 250 kVA.
- Cable troncal de aluminio con una sección de 15 mm$^2$ para observar más claramente el efecto de la caída de tensión.
- Cable de acometida de cobre con una sección transversal de 15 mm$^2$.
- La distancia entre el transformador y el primer cliente conectado es de 20 metros.
- Los clientes se ubican equidistantes entre sí, con una distancia de separación de 10 metros.
- Se consideran 60 clientes en total, una cantidad aproximada que se encuentra en circuitos en zonas urbanas alimentadas por transformadores de 250 kVA donde generalmente se encuentran tres o más circuitos.
- Inicialmente, se asume un valor constante para la potencia activa y reactiva de todos los clientes. El valor tomado es de 200 W para la potencia activa, y un 10 % de este valor para la potencia reactiva. Estos valores, si bien son más bajos que los que se podrían encontrar en barrios residenciales en Tucumán, son adecuados como caso teórico de prueba.

## Esquema de conexión de fases 112233

Para evaluar diferentes aspectos del método de balance de cargas propuesto, se crearon varios casos de prueba seleccionando las fases de conexión de cada cliente según un esquema particular. El esquema *112233* consiste en dividir a los clientes en 3 grupos. El primer tercio de clientes, ordenados por su distancia al transformador, se asigna a la fase 1; el segundo tercio se asigna a la fase 2; y el último tercio se asigna a la fase 3. La Figura 5 muestra una red eléctrica simple con este esquema de conexión. Con este caso se desea evaluar principalmente el impacto de la caída de tensión, dado que el desbalance de consumo es cercano a cero porque la cantidad de clientes por fase es la misma y por lo tanto el consumo total por fase es aproximadamente igual. Este esquema genera una caída de tensión desigual en cada fase, ya que la distancia desde el transformador hasta el último cliente varía entre fases. Es decir, para los clientes en la fase 3, se requiere un cable de mayor extensión, lo que conlleva una mayor caída de tensión. Dado que todos los clientes demandan la misma cantidad de corriente, la caída de tensión depende principalmente de la distancia a la que está conectado el cliente.

La Tabla 1 muestra los valores obtenidos simulando la red eléctrica con este esquema de conexión de fases (columna "Inicial"). El desbalance es muy bajo como era de esperarse, rondando el 3.4 % muy por debajo del límite establecido en 15 %. Sin embargo, la caída de tensión en la fase 3 es de 7.8 %, la cual está por encima del valor deseado de 5 %. Esto se ve reflejado en la Figura 6 que muestra cómo cae la tensión a lo largo del circuito.

En primer lugar se buscará todos los cambios de fase que fuera necesarios para minimizar el desbalance de carga como único objetivo de la optimización. De esta manera se evaluará cómo se comporta el algoritmo genético en cuanto a la cantidad total de cambios requeridos y cómo se ve afectada la caída de tensión a pesar de no incluir estos aspectos como objetivo de la búsqueda. Los valores de los parámetros utilizados fueron $\alpha = 1{,}0$ (desbalance), $\beta = 0{,}0$ (caída de tensión), $\gamma = 0{,}0$ (cantidad de cambios). Luego, se hizo una prueba considerando sólo la caída de tensión en la optimización para determinar cómo se comporta el algoritmo genético en cuanto a si la reducción de la caída de tensión tiene influencia directa en el balance de carga o son objetivos contrapuestos. Se utilizaron los parámetros $\alpha = 0{,}0$ (desbalance), $\beta = 1{,}0$ (caída de tensión), $\gamma = 0{,}0$ (cantidad de cambios). La siguiente prueba propone evaluar el comportamiento del algoritmo cuando se consideran dos objetivos: minimizar el desbalance de carga y minimizar la caída de tensión en el fin de línea. Se asigna $\alpha = 1{,}0$ (desbalance), $\beta = 1{,}0$ (caída de tensión), $\gamma = 0{,}0$ (cantidad de cambios) como pesos de cada componente de la función fitness. Finalmente, se combinan todos los objetivos incluyendo el componente que controla la cantidad de cambios de fase a realizar. En las ejecuciones anteriores no se incluyó la componente de cantidad de cambios en la función objetivo. Entonces, se daba libertad al algoritmo de encontrar soluciones que cumplan con el objetivo planteado sin importar cuántos clientes deban cambiar de fase. Sin embargo, una gran cantidad de cambios no es deseable en la práctica porque involucra altos costos por el envío de cuadrillas, y molestias a los clientes por requerir suspender momentáneamente el suministro. Para habilitar esta restricción se utilizarán los siguientes pesos de cada componente de la función fitness son: $\alpha = 1{,}0$ (desbalance), $\beta = 1{,}0$ (caída de tensión), $\gamma = 0{,}6$ (cantidad de cambios).

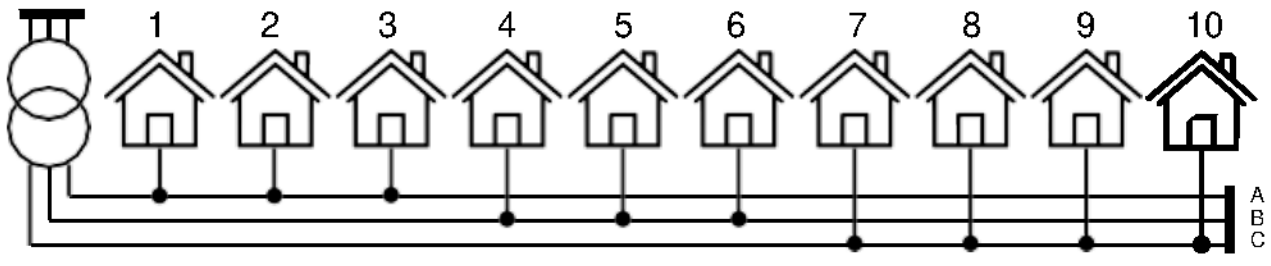

**Figura 5: Modelo de red eléctrica simple usando esquema de conexiones de fase 112233.**

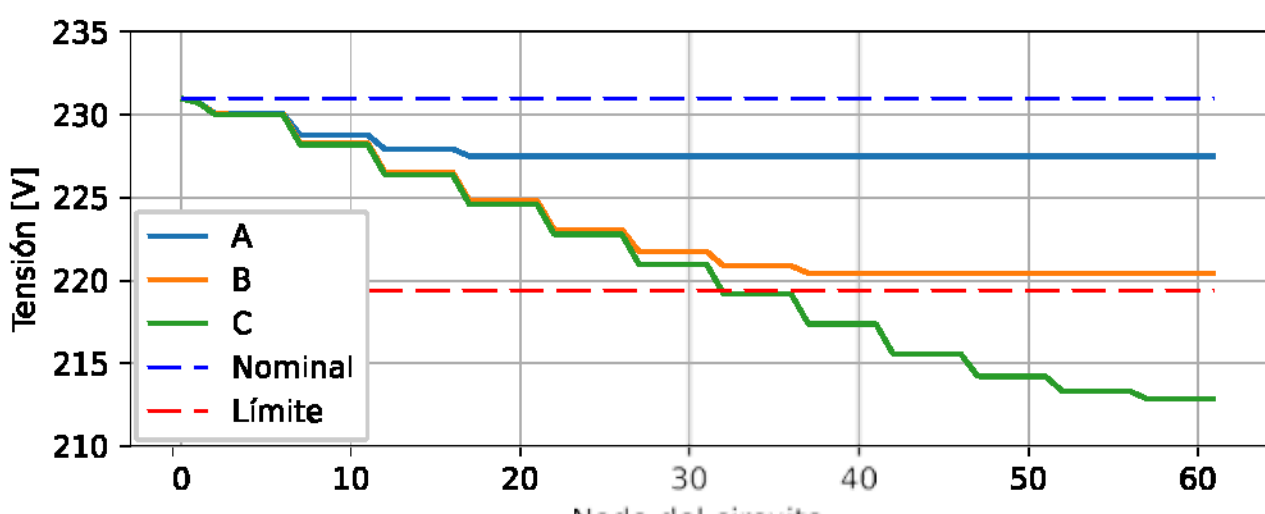

**Figura 6: Caída de tensión en el circuito con conexión de fases con esquema 112233.**

**Sólo desbalance:** Los valores finales se muestran en la columna "Desbalance" de la Tabla 1. El valor de desbalance final es bajo, rondando el 1.32 %. Esta solución se considera óptima debido a que cada fase tiene una pequeña corriente asociada a pérdidas técnicas distinta en cada cable y que no puede eliminarse. Por otro lado, la caída de tensión más alta se observa en la fase 3, con un valor de 5.99 %, que es menor que el valor antes del cambio. Esta mejora no es resultado de la optimización, sino una coincidencia. Esto demuestra la necesidad de incluir la componente de caída de tensión en fin de línea en la función objetivo para lograr una optimización adecuada. Finalmente, en cuanto a la cantidad de clientes a cambiar, el total es de 18, aunque no sería necesario realizar tantos cambios para equilibrar el desbalance, ya que la demanda total de energía por fase era la misma antes de la optimización.

**Sólo caída de tensión:** La columna "Tensión" de la Tabla 1 muestra los resultados obtenidos. En este caso, se observa una mejora significativa en la caída de tensión en el fin de línea, la cual se incrementa progresivamente hasta alcanzar valores aceptables menores al 4.62 % en todas las fases. Sin embargo, aunque se cumple el objetivo de optimización en cuanto a la caída de tensión, el desbalance de carga ha empeorado respecto a la situación inicial, con un índice de 39.1 %, superando el valor deseado del 15 %. Además, la cantidad de cambios requeridos por esta solución es de 18, lo cual es excesivo considerando que se podrían obtener resultados similares moviendo menos clientes de la fase más problemática. Estos resultados indican la necesidad de incluir otros componentes en la función de fitness para lograr soluciones más equilibradas y prácticas.

**Desbalance y caída de tensión:** A diferencia de la ejecución anterior, el algoritmo requiere más generaciones para estabilizarse. Los resultados obtenidos se muestran en la columna "Desb. + Tensión" de la Tabla 1. Siempre habrá una caída de tensión en el fin de línea y una corriente que genera desbalance en el sistema, incluso si la cantidad de clientes por fase es la misma y todos consumen la misma cantidad de energía. En este caso, el desbalance de carga es cercano a 0 debido a que la cantidad de clientes conectados en cada fase es igual y se ha asumido que todos consumen lo mismo, resultando en un desbalance de carga cercano a cero. La caída de tensión en la fase con menor tensión es de 4.62 %, cumpliendo con el valor máximo deseado del 5 %. Sin embargo, la solución encontrada requiere cambiar de fase a 26 clientes, es decir, el 43 % de los clientes en la red. Esto indica que, para cumplir con los objetivos establecidos, el algoritmo debe reconectar un número considerable de clientes. Esto resalta la necesidad de ajustar la búsqueda para encontrar soluciones que impliquen un menor número de cambios.

**Tabla 1: Valores de corriente y tensión obtenidos a partir de simulación de red eléctrica con esquema de conexión de fase 112233.**

| | Inicial | | | Desbalance | | | Tensión | | | Desb. + Tensión | | | Desb. + Tensión + Cambios | | |
|---|---|---|---|---|---|---|---|---|---|---|---|---|---|---|---|
| Fase | A | B | C | A | B | C | A | B | C | A | B | C | A | B | C |
| Tensión en fin de línea [V] | 227.5 | 220.4 | 212.8 | 223.4 | 220.7 | 217.1 | 220.7 | 220.4 | 220.3 | 220.5 | 220.6 | 220.3 | 221.2 | 220.4 | 219.7 |
| Caída de tensión [ %] | 1.49 | 4.56 | 7.84 | 3.26 | 4.43 | 5.99 | 4.42 | 4.58 | 4.62 | 4.54 | 4.47 | 4.62 | 4.22 | 4.56 | 4.87 |
| Corriente por fase [A] | 17.6 | 18.6 | 18.9 | 17.7 | 18.0 | 18.3 | 25.0 | 16.3 | 12.7 | 17.9 | 18.0 | 17.9 | 17.9 | 18.1 | 18.0 |
| Cantidad de clientes | 20 | 20 | 20 | 20 | 20 | 20 | 28 | 18 | 14 | 20 | 20 | 20 | 20 | 20 | 20 |
| Desbalance de carga [ %] | 3.37 | | | 1.32 | | | 39.1 | | | 0.38 | | | 0.77 | | |
| Cantidad de cambios | - | | | 18 | | | 18 | | | 26 | | | 14 | | |

**Todos los objetivos combinados**: La mejor solución encontrada tiene los valores finales reportados en la columna "Desb. + Tensión + Cambios" de la Tabla 1. El desbalance de carga es menor a 1.0 % y la tensión cae hasta 4.9 %, ambos dentro del rango deseado. Esto se ve reflejado en la Figura 8 donde se observa la caída de tensión a lo largo del circuito en las 3 fases. La cantidad de cambios es 14, menor que la ejecución en el caso anterior donde no fue incluida en la función de fitness. Esto comprueba que se puede obtener similares resultados pero con una menor cantidad de cambios. Si vemos la evolución de cada objetivo por separado en la Figura 7 se puede observar que todos ellos se estabilizan en las últimas generaciones hasta alcanzar su valor final.

La solución encontrada cumple con los criterios establecidos. Los valores de los pesos elegidos para cada objetivo dio como resultado una solución que mejora en todos los aspectos la situación de la red eléctrica con relativamente pocos cambios necesarios. Si se quisiera reducir aún más la cantidad de clientes necesarios a cambiar se podría aumentar el valor de $\gamma$ a costa de obtener soluciones con un poco más de desbalance y caída de tensión.

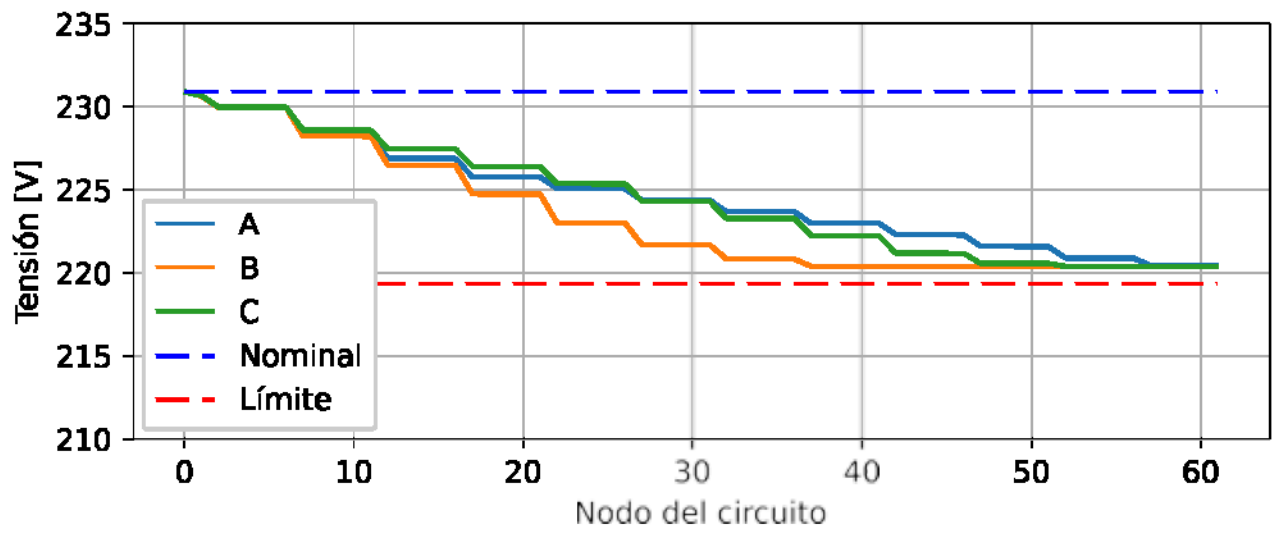

**Figura 8: Caída de tensión en el circuito con esquema de conexión de fases 112233 después de realizados los cambios de fase seleccionados.**

## Otros esquemas de conexión

**Esquema de conexión de fases *123123***: asigna alternadamente la fase 1, 2 o 3 a lo largo del circuito, tal como muestra el ejemplo de la Figura 9a. Este ejemplo teórico representa el caso más favorable, ya que debido a que los clientes están distribuidos uniformemente a lo largo del circuito y que todos los clientes tienen el mismo valor de consumo, entonces el desbalance de carga es cercano a cero y la caída de tensión a lo largo de la línea es igual para cada fase. Por lo tanto, para este caso no es necesario aplicar el método de balance de carga, pero lo haremos para mostrar que nuestro método reconoce este tipo de situaciones y propone muy pocos cambios.

La Tabla 2 muestra los valores resultantes. Como era de esperarse, el índice de desbalance es aproximadamente 0 debido a las características del caso, la caída de tensión

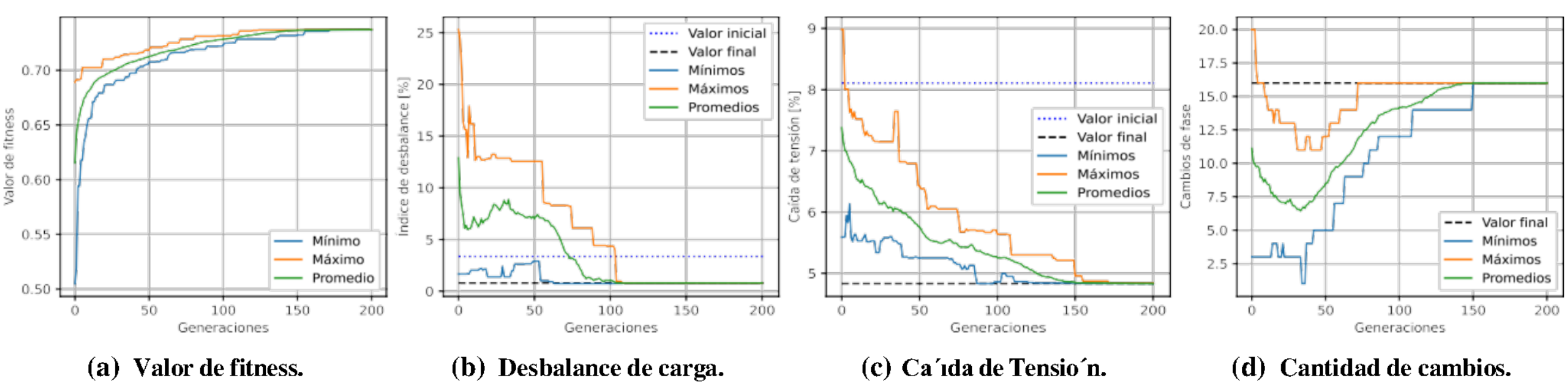

**(a) Valor de fitness.** **(b) Desbalance de carga.** **(c) Caída de Tensión.** **(d) Cantidad de cambios.**

**Figura 7: Evolución de valores de cada objetivo independiente a través de las generaciones en una ejecución del Algoritmo Genético para balance de carga, en un esquema de conexión de fases 112233.**

**Tabla 2:** **Valores de corriente y tensión obtenidos a partir de simulación de red eléctrica con otros esquemas de conexión de fase.**

| | Esquema de conexión 123123 | | | | | | Esquema de conexión 111 | | | | | |
|---|---|---|---|---|---|---|---|---|---|---|---|---|
| | Antes | | | Después | | | Antes | | | Después | | |
| Fase | A | B | C | A | B | C | A | B | C | A | B | C |
| Tensión en fin de línea [V] | 220.8 | 220.5 | 220.1 | 220.4 | 220.5 | 220.5 | 196.3 | 230.9 | 230.9 | 220.4 | 220.6 | 220.4 |
| Caída de tensión [ %] | 4.38 | 4.54 | 4.69 | 4.54 | 4.54 | 4.54 | 15.00 | 0.00 | 0.00 | 4.58 | 4.46 | 4.58 |
| Corriente por fase [A] | 17.9 | 18.0 | 18.0 | 18.0 | 18.0 | 20.0 | 58.2 | 0.0 | 0.0 | 18.0 | 18.0 | 18.0 |
| Cantidad de clientes | 20 | 20 | 20 | 20 | 20 | 20 | 60 | 0 | 0 | 20 | 20 | 20 |
| Desbalance de carga [ %] | 0.16 | | | 0.06 | | | 200.0 | | | 0.09 | | |
| Cantidad de cambios | - | | | 2 | | | - | | | 40 | | |

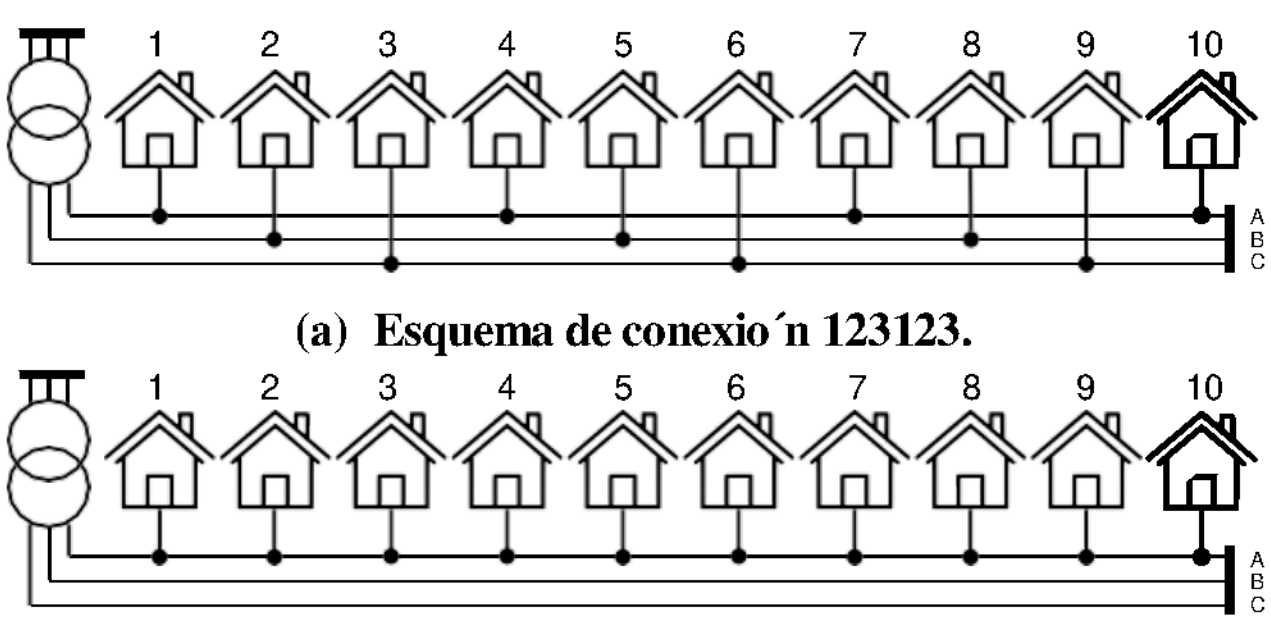

(a) **Esquema de conexión 123123.**

(b) **Esquema de conexión 111.**

**Figura 9:** **Modelo de red eléctrica simple utilizando otros esquemas de conexiones de fase.**

es de aproximadamente 5 %, lo que está justo sobre el valor umbral deseado, y la cantidad de cambios es pequeña.

**Esquema de conexión *111***: simplemente asigna la misma fase 1 a todos los clientes de la red. La Figura 9b muestra este esquema mediante un ejemplo sencillo con 10 clientes. Este es el caso más desfavorable tanto para el índice de desbalance como para la caída de tensión de fin de línea. El desbalance de carga tiene un índice con valor más grande debido a que la diferencia entre el consumo de la primera fase respecto a las otras es máxima. Lo mismo sucede con la caída de tensión, ya que la tensión en la fase más cargada disminuye a lo largo de toda la línea. Además, teniendo en cuenta que cada cliente consume la misma cantidad de energía, una buena solución debería equilibrar la cantidad de clientes en cada fase. La cantidad máxima de cambios deberían ser 40 (dos tercios de la cantidad total de clientes $N = 60$). Por estas características se toma este caso de prueba como base para las pruebas preliminares.

La Tabla 2 muestra los valores obtenidos simulando la red eléctrica con este esquema de conexión de fases antes de realizar cualquier cambio. El índice de desbalance es el más alto posible igual a 200 %, es decir que la corriente de la fase más alta es 2 veces más grande que la corriente promedio (valor que tendría el sistema perfectamente balanceado). La caída de tensión en la fase 1 es 15 %, muy por debajo de los valores tolerables para los equipos eléctricos que se alimentan de la red.

## Conclusiones

En este trabajo se propuso un método para realizar el balance de cargas en las fases del transformador. El mismo está basado en Algoritmos Genéticos para encontrar la fase adecuada de un porcentaje de los clientes de la red de los cuales se dispone de mediciones del consumo eléctrico. Está diseñado considerando que los cambios de fase propuestos serán realizados manualmente, a través de un operario especializado. Por lo tanto, además de reducir el desbalance de cargas en las fases del transformador durante un periodo, en la búsqueda se tiene en cuenta minimizar la caída de tensión en el último nodo de la red, mientras que también se busca minimizar la cantidad de cambios requeridos y que la nueva configuración de fases se mantenga en el tiempo.

Se propone como trabajo futuro evaluar el algoritmo desarrollado en este trabajo en casos reales, modelando topologías de redes más complejas. Se analizarán otras maneras de definir la función de evaluación, con el objetivo de facilitar aún más el ajuste de parámetros y permitir el uso de otro tipo de algoritmos de búsqueda. Finalmente, se evaluará la duración de las soluciones encontradas por la heurística, es decir, determinar durante cuánto tiempo se mantienen balanceadas las cargas luego de aplicados los cambios.

## Agradecimientos

Este trabajo fue soportado parcialmente por el subsidio asignado al proyecto UTN "Análisis Inteligente de Datos Aplicado a Gestión y Optimización de la Energía II" (CCUTITU7896TC). Agradecemos a la Empresa de Distribución de Energía de Tucumán (EDET S.A.) por proporcionar los datos necesarios para la elaboración de este trabajo, previstos en el marco del convenio con el grupo GITIA en 2022.

## Referencias